\documentclass[letterpaper, 10 pt, conference]{ieeeconf}  

\IEEEoverridecommandlockouts                              

\usepackage{cite}
\usepackage{amsmath,amssymb,amsfonts}
\usepackage{algorithmic}
\usepackage{graphicx}
\usepackage{textcomp}
\usepackage{float}
\usepackage{pifont}
\usepackage{array}
\usepackage{multirow}
\usepackage{multicol}
\usepackage[dvipsnames,table,xcdraw]{xcolor}
\usepackage{subcaption}
\usepackage{dsfont}
\newcommand{\mC}{{\color{OliveGreen}\ding{51}}}
\newcommand{\mX}{{\color{BrickRed}\ding{53}}}
\usepackage{tabularx}

\title{\LARGE \bf A model-free approach to fingertip slip and disturbance detection for grasp stability inference
}

\author{Dounia Kitouni*, Mahdi Khoramshahi and Veronique Perdereau 
\thanks{All authors are with the Institute of Robotics and Intelligent Systems, 
        Sorbonne University, Paris, France}
\thanks{* Corresponding author e-mail: {\tt\small kitouni@isir.upmc.fr}}
\thanks{This work was supported by CORSMAL (CHIST-ERA program) under grant EP/S031715/1}%
}

\begin{document}

\maketitle
\thispagestyle{empty}
\pagestyle{empty}

\begin{abstract}
Robotic capacities in object manipulation are incomparable to those of humans. Besides years of learning, humans rely heavily on the richness of information from physical interaction with the environment.
In particular, tactile sensing is crucial in providing such rich feedback.
Despite its potential contributions to robotic manipulation, tactile sensing is less exploited; mainly due to the complexity of the time series provided by tactile sensors.
In this work, we propose a method for assessing grasp stability using tactile sensing.
More specifically, we propose a methodology to extract task-relevant features and design efficient classifiers to detect object slippage with respect to individual fingertips.
We compare two classification models: support vector machine and logistic regression.
We use highly sensitive Uskin tactile sensors mounted on an Allegro hand to test and validate our method. 
Our results demonstrate that the proposed method is effective in slippage detection in an online fashion.
\end{abstract}

\section{Introduction}
In dexterous object manipulation, grasp stability is the problem that deals with locking the object and rejecting external forces.
Solving this problem is complex and depends on hand kinematics, interaction forces with the object, and environmental uncertainties.
The solution is even more challenging if object characteristics, e.g., shape, weight, stiffness, or friction properties are unknown. 
Approaches using compliant under-actuated hands adopt power grasps and leverage the hand's adaptability to cope with object uncertainties.
While this simplifies the grasp stability problem, it is limited to pick and place tasks and not suitable for dexterous manipulation \cite{chen2020analysis,kiatos2019grasping,zeng2021learning}.
On the other hand, precision grasps prioritize dexterity over stability \cite{stachowsky2016slip}.
Other approaches to dexterous manipulation make several assumptions about the object and interaction dynamics. which allows offline grasp planning \cite{calandra2018more,merzic2019leveraging,brahmbhatt2019contactgrasp,datadriven2013}.
However, when these planned grasps are executed, they are affected by real-world uncertainties and model inaccuracies, leading to slippage and loss of contact\cite{datadriven2013}.
During dexterous manipulation, humans rely on their sense of touch to gain essential information about physical interaction with objects.
Researchers in robotics mimic this ability by using tactile sensors and implementing slip-detection algorithms to detect object movements relative to the fingers\cite{liu2020bioinspired}.
The outputs of these algorithms can be used to design reactive grasping strategies robust to the aforementioned uncertainties.

\begin{figure}[t]
  \centering
  \begin{subfigure}{0.49\linewidth}
    \centering
    \includegraphics[width=\linewidth]{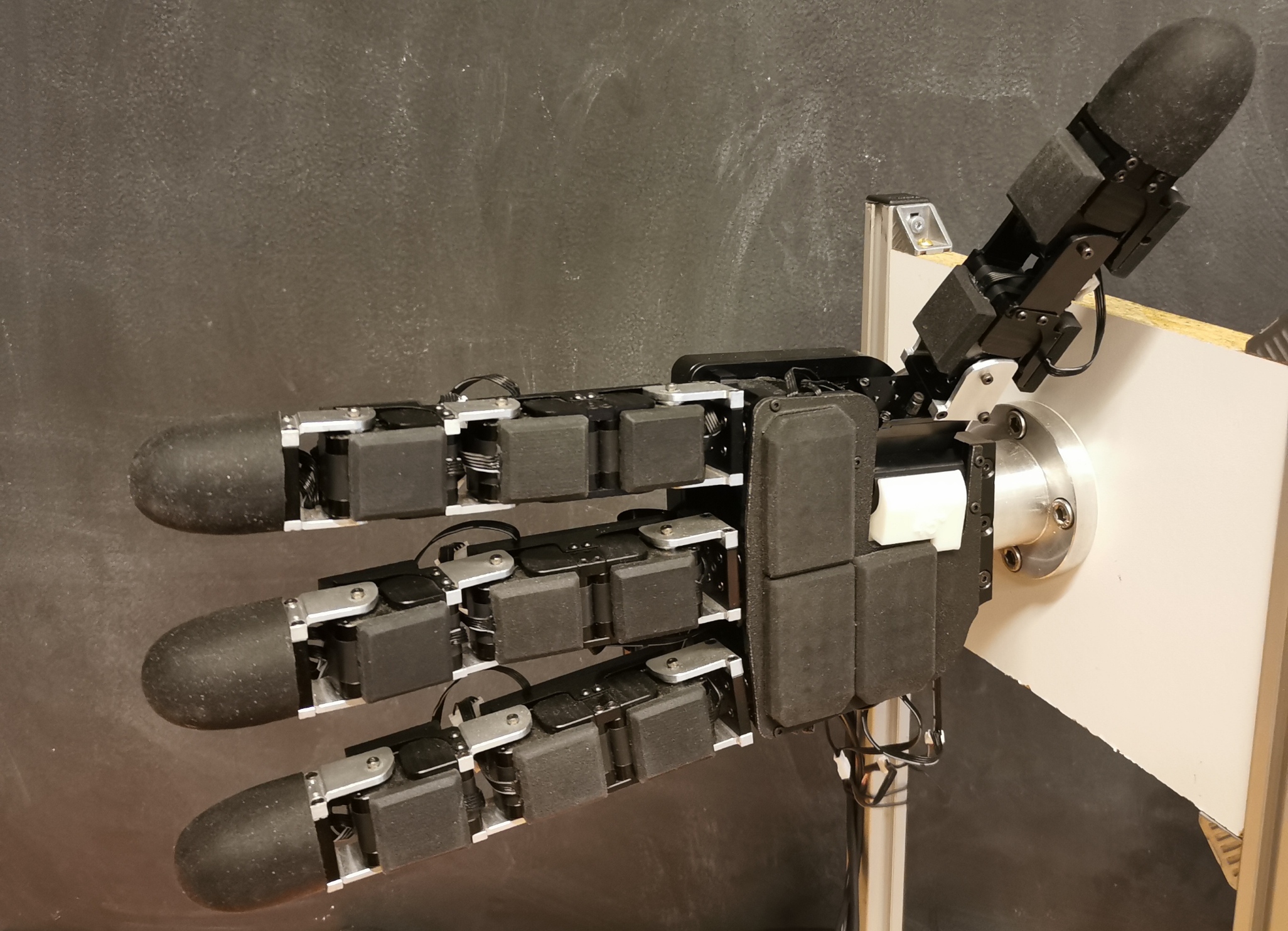}
   \caption{\footnotesize Allegro hand with tactile sensors}
    \label{fig:f3}
  \end{subfigure}
  \hfill
  \begin{subfigure}{0.49\linewidth}
    \centering
    \includegraphics[width=\linewidth]{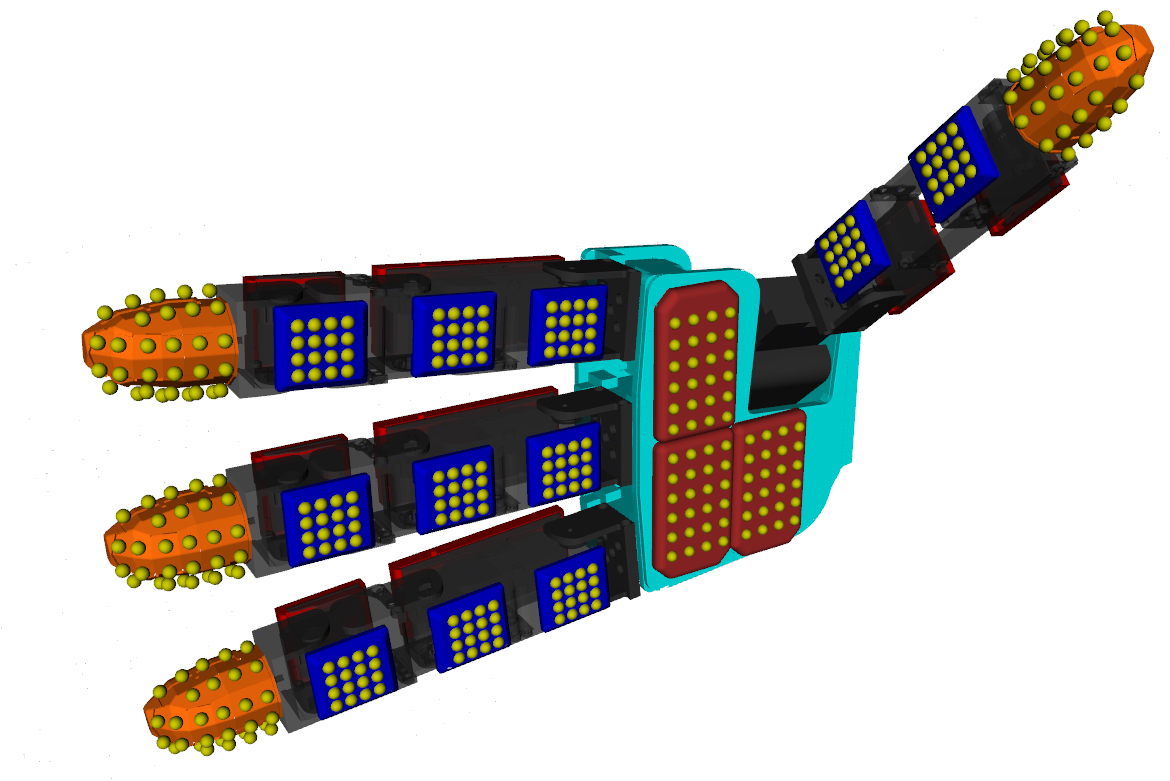}
    \caption{\footnotesize 3D model with taxels positions}
    \label{fig:f4}
 \end{subfigure}
  \caption{The Allegro hand with tactile sensors and a 3D model view to show taxels positions.}
  \label{fig:allegro_hand}
\end{figure}

In this work, we contribute to the literature on  assessing grasp stability by proposing efficient classifiers for slip detection where Discrete Wavelet transform DWT is used for feature extraction. 
We classify the contacts with the object into ``stable'' and ``unstable''.
The ``unstable'' class includes both slippage and loss of contact between the fingertip and the object.
Our method is robust against the magnitude of the interaction forces, i.e., it gives a satisfactory detection rate when the applied forces are either high or low.
Additionally, our method provides a measure of instability for each fingertip, which is beneficial to designing independent feedback control strategies, e.g., during finger-gaiting and in-hand manipulation tasks.
To validate our method, we use a multi-fingered hand equipped with 3-axial tactile sensors to perform precision grasps around rigid objects (Fig.~\ref{fig:allegro_hand}).

\section{Methods for slip detection}
In a recent review, Romero et al.~\cite{romeo2021method} broadly categorized slip detection methods using tactile sensing into four groups: friction-based, vibration-based, differentiation-based, and learning-based. They also explain that the choice of the method is highly dependent on the output of the tactile.
\textbf{Friction-based} methods are typically used when 3-axial force information is available and rely on the static friction coefficient between normal and tangential forces \cite{melchiorri2000slip,okatani2017mems, beccai2008development,zhang2014multifingered,kaboli2016tactile}. These methods have several advantages, including fast prediction/detection and noise robustness.
However, they heavily depend on complex contact models with numerous parameters (such as the Coulomb contact model), limiting their real-world application.
\textbf{Vibration-based} methods aim to detect the mechanical vibrations caused by a slip in a model-free manner \cite{fernandez2014micro,damian2010artificial,cheng2015data, romeo2017slippage}. They are used when only the normal component of the force is available.
Researchers have proposed various approaches to detect slips based on measuring high-frequency components of the output, most often based on Fast Fourier Transform (FFT) \cite{romano2011human}. 
However, using FFT results in the loss of temporal information. 
To address this limitation, researchers use Discrete Wavelet Transform (DWT) \cite{yang2015new,wang2016slip,deng2016wavelet,romeo2018identification} over short time windows to extract high-frequency components to detect slip. 
Nevertheless,  DWT methods introduce a delay in the signal, increasing the rate of false positives.
\textbf{Differentiation-based} methods \cite{lee2018development,feng2019slip, osborn2013utilizing} detect slip by observing abrupt changes in the sensor's output. They are typically used when only the normal component of the force is available. 
These methods are highly slip-sensitive and do not require any friction model or frequency analysis. 
However, they have a high rate of false positives.  
\textbf{Learning-based} methods \cite{james2020slip,agriomallos2018slippage,zapata2019tactile,li2018slip,muthusamy2020neuromorphic} utilize machine learning techniques (such as neural networks) 
to learn slip models from a labeled dataset of tactile sensor output. They are commonly used when the output of tactile sensors is in the form of images or patterns, and the other previously mentioned methods can't be used. They are also used when only the normal force is available or all its components are.
Learning-based methods formulate slip detection as a classification problem where the data points are labeled ``slip'' and ``no-slip''.
Robustness to noise and handling non-linearities are the main advantages of these methods.
However, they require many labeled data and a long training process.
Moreover, implementing and integrating these methods with real-world robotic systems is not straightforward.

In this work, we combine vibration methods with learning methods.  We use DWT to retrieve temporal information from the tactile signal, which is crucial for slip detection.
We then extract features from the transform, and we use two different classifiers to detect not only slips but also instabilities at the finger level.
\section{Problem description}
Considering tactile feedback, we build a system to independently predict instabilities due to slippage and disturbances for fingertips without prior knowledge about the object proprieties and without making assumptions about contact models. In this work, we consider that instabilities can take two forms.
The first one is when the hand is closing on the object before it is locked in the absence of external forces.
The second is when external forces are being applied to the object; we will refer to them as disturbances resulting in slippage or loss of contact, or both.  We aim to provide measures for this instability that the controller can exploit.

To test our method, we use the Allegro hand from Wonik Robotics. 
It is a low-cost hand with four fingers and sixteen independent torque-controlled joints. 
We equipped the hand with Uskin soft sensors \cite{tomo2017covering} covering the palm, the phalanges, and the fingertips' curved surface with a total of 368 taxels Fig.~\ref{fig:allegro_hand}. 
Uskin soft sensors use magnetic field changes induced by the skin deformation to provide 3-axial force measurements.
The contact geometry of the letter influences the measurements; thus, the sensors will be uncalibrated in this work.
\section{Method description}
\subsection{Tactile data pre-processing } \label{preprocessing}
\begin{figure}[h]
    \centering
    \includegraphics[width=0.7\linewidth]{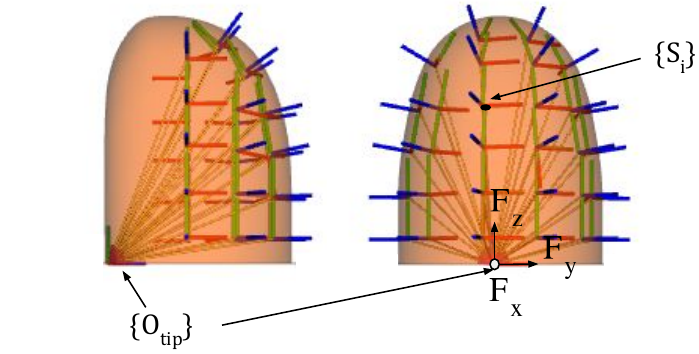}
    \caption{Side and front view of 3D Uskin sensor at the fingertip, shows the placement of each taxel and its corresponding reference frame $\{S_i\}$ in relation to the origin of the sensor.}
    \label{fig:fingertiptaxls}
\end{figure}

\begin{figure*}[t]
    \centering
    \includegraphics[width=\linewidth]{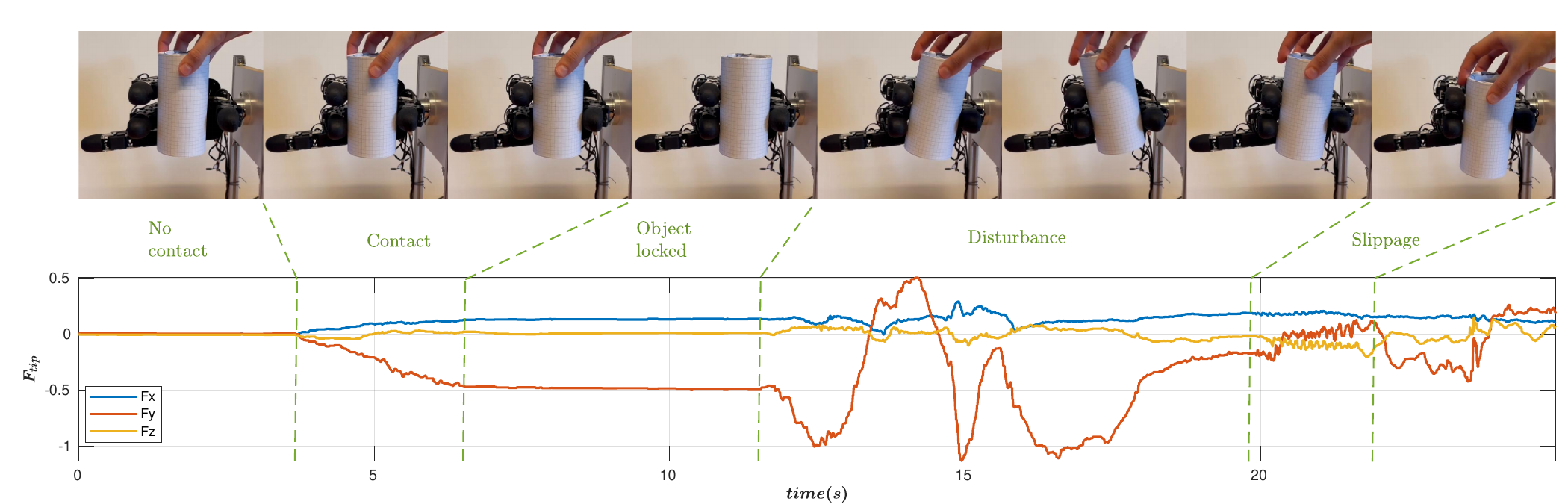}
    \caption{In this figure, the human hands over the object while the robot closes its fingers. \textbf{No contact}: Before the fingertips reach the object's surface, the force equals zero. \textbf{Contact}: As the contact starts, the force grows to lock the object. \textbf{Object locked:} When the object is locked, the force amount is constant. \textbf{Disturbances}  We manually move the object to mimic disturbances caused by external forces. \textbf{Slippage} we also push downward to simulate slippage.}
    \label{fig:pics_force}
\end{figure*}

Each fingertip has a total of $N_{s}=30$ taxel sensors distributed along the known surface of the fingertip. 
For time-step $k$, taxel $i$ provides the measured force $f_i(k) \in \mathds{R}^3$, which is specified in the reference frame placed at the origin of the taxel  $ \{S_i\} $  as shown in  Fig.~\ref{fig:fingertiptaxls}.
Let $\{O_{tip}\}$ be the origin of an arbitrary frame attached to the last segment of the finger with the known rotation matrix $R_i 
\in SO(3)$ for each $ \{S_i\} $.
We compute the total force applied to each fingertip as follows:
$$F_{tip}(k)=\sum_{i=1}^{N_{s}} R_i f_i(k)$$
with $ F_{tip}(k)=[F_x,F_y,F_z] \in \mathds{R}^3$. Fig.~\ref{fig:pics_force} shows the three components of $F_{tip}$. 
\subsection{Features extraction }\label{Featuresextraction}
In this part, we detail how we extract the relevant temporal information from force measurements. 
First, we compute $F_a(k)=|{F}_{tip}(k)|$ as the amplitude of the fingertip force at time-step $k$.
Discrete wavelet transform decomposes the signal into approximations $A(k)$ and details $ D(k)$ as so: $F_a(k) =A(k)+D(k)$

 \noindent with:
 {\small
 \begin{align*}
     A(k)&=\sum_{m=k-N_w}^{k} a_{\phi}(k) \phi(k-m)\\
     D(k)&=\sum_{m=k-N_w}^{k} d_{\psi}(k) \psi(k-m)
 \end{align*}}
 
\noindent where $\phi(.)$ and $\psi(.)$ are the scaling function and the mother wavelet, respectively. 
Moreover, $a_{\phi}(k)$ and $d_{\psi}(k) \in \mathds{R}$ are the approximation coefficients and the detail coefficients respectively. $N_w$ is a positive even number representing the window size we will use to compute the DWT transform. $n$ is the time translation \cite{gao2010wavelets}.

In this work, we use the Haar wavelet transform (Haar coefficients) since it can be simply implanted in real-time and is computationally cheap. 
In the resulting decomposition, the approximation coefficients $a_{\phi}$ carry information about the characteristics of the signal in low frequencies, whereas the detail coefficients $d_{\psi}$ carry information about the characteristics of the signal in high frequencies.

When disturbances occur, they appear in the signal as abrupt changes in the amplitude of the force, that is, high frequencies. The DWT on small time windows allows us to capture these changes.

We use the extracted coefficients to compute the following features.
We compute the moving-average of $a_{\phi}$ as:
\begin{equation}
    m(k)= \frac{1}{N_w}\sum{a_{\phi}(k)}
\end{equation}
Similarly, we define $\sigma$ the standard deviation of $d_{\psi}$ as:
\begin{equation}
    \sigma(k)= \sqrt{\sum {\frac{(d_{\psi}(k)-\Bar{d_{\psi}})^2}{N_w}}}
\end{equation}
The moving average on the approximation coefficients reduces the noise due to measurement since no previous filtering of the sensors has been done. Hence, it gives more accurate information about the amplitude of the force. On the other hand, the standard deviation measures fluctuations in the details caused by abrupt changes in the force amplitude.\\
For every time step $k$ we construct a feature vector $\Phi(k)$ as:

$$ \Phi(k) =[ F_a(k), F_{tip}(k), m(k), \sigma(k) ]^T $$

\subsection{Data collection and labeling}
\begin{figure}[b]
    \centering
    \includegraphics[scale=0.047]{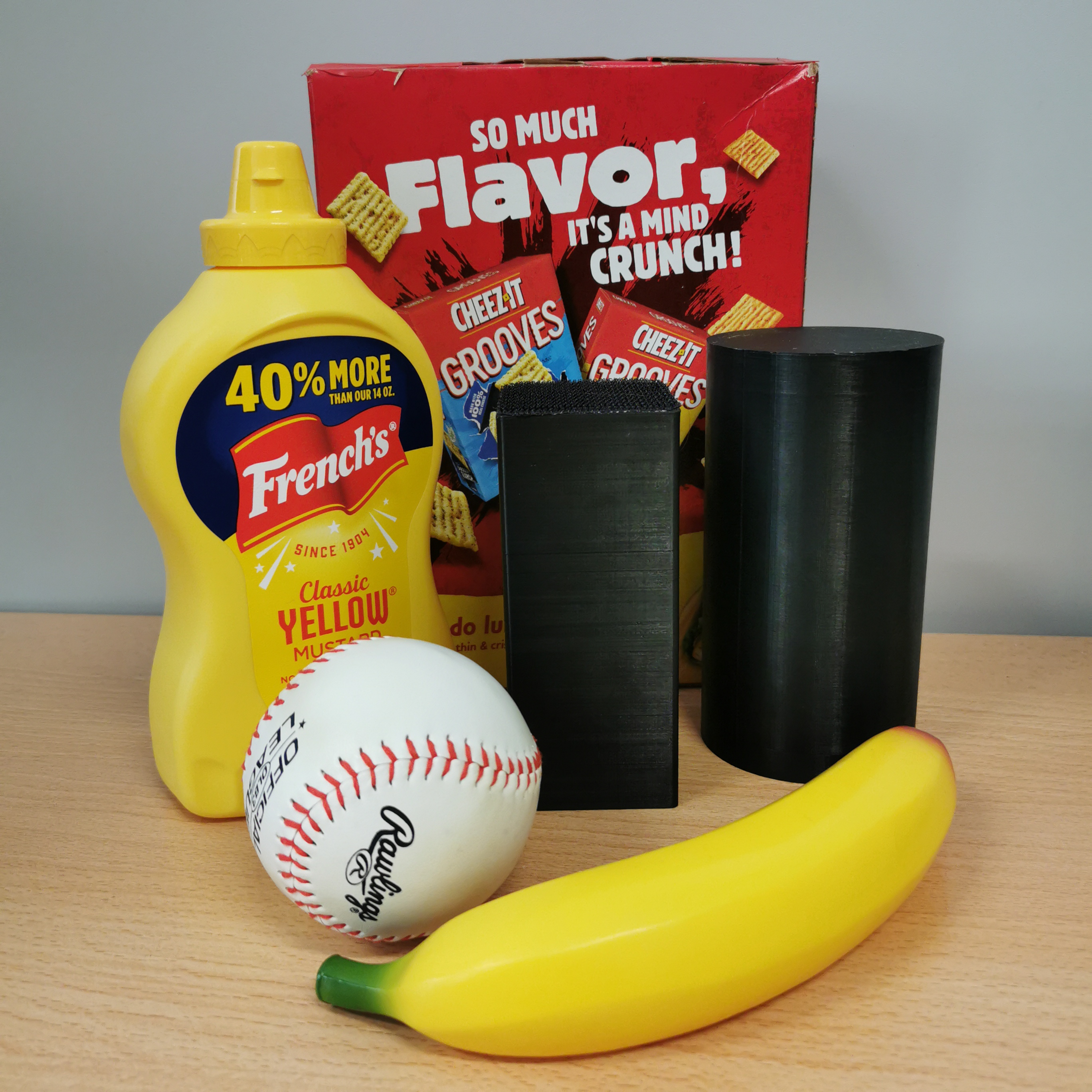}
    \caption{The objects used for training and testing the classifiers. The objects have different characteristics: different shapes, textures, and weights }
    \label{fig:obj}
\end{figure}

To collect data for this work, we will consider two cases. In the first case, we will place an object in the hand's workspace. 
We then close the fingers around a predefined axis until the hand fully locks the object.
We maintain the grasp for at least 10 seconds, and we then release it.
In the second case, we repeat the same steps as the first case. When the hand fully locks the object, we manually apply external forces to create disturbances. We push the object downward/upward, left/right, and randomly.
A total of 6 objects were used to collect data for the training Fig.~\ref{fig:obj}, The objects have different weights and textures.
For each object, the two cases were repeated 5 times.

Fig.~\ref{fig:pics_force} illustrates this procedure for one experiment. 
A total of 250000 data was collected, corresponding to all experiments for one fingertip.
The data was recorded with a frequency of 150 data per second.
We extracted features explained in sections \ref{preprocessing} and \ref{Featuresextraction}.
 To construct the features vector $\Phi$, we used a sliding window size of $N_w=14$ for the Haar wavelet decomposition. 80\% of the data were used for training the two classifiers. The rest 20\% were used for testing.
 
\begin{figure}[!ht]
    \centering
    \includegraphics[width=0.9\linewidth]{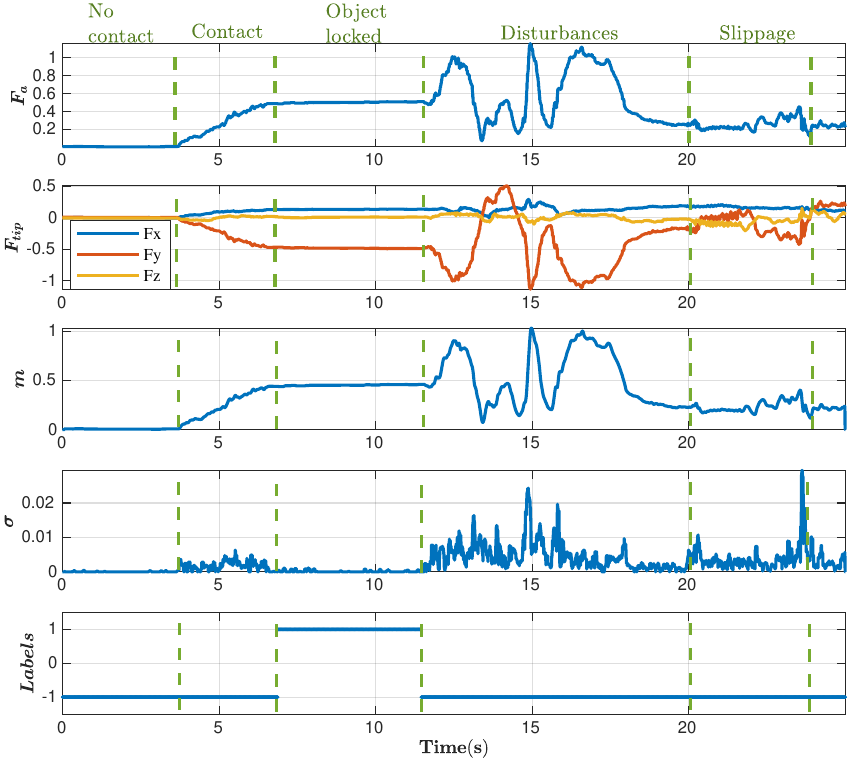}
    \caption{Extracted features and labels for a single fingertip obtained from an experiment involving a human handing over an object to a robot, which locks the object, and then disturbances are manually applied. Stable features (obtained when the object is locked) are labeled as $1$, while unstable features are labeled as $0$.}
    \label{fig:expiriment}
\end{figure}

 Fig.\ref{fig:expiriment} corresponds to the data collected in one experiment and the features extracted from it.
In this figure, the total force equals zero as the hand closes and before the fingers reach the object. 
When the contact occurs, the force increases to lock the object. When the object is locked, the force amplitude is constant. Then comes the disturbances. We label the features when the object is locked as $1$, corresponding to stability. The rest is labeled as $0$. It comprises data in the absence of contact, the features preceding the stable contact when the force component is growing, and the features corresponding to the disturbances.
\subsection{Model selection}
Our goal in this part is to separate stable contacts from unstable contacts.
By stable, we refer to the state when the contact is maintained on the object. The object's dynamics are fully governed by the robotic hand. Unstable contacts include the absence of contact, disturbances, and slippage. The absence of contact can be separated from the unstable contact by using a small threshold on the amount of the amplitude of the measured force.
When using the classifier, we aim to assign the value $1$ to the feature vector $\Phi$ when the contact is stable and $0$ otherwise. For the classification, we use two different methods: Support Vector Machines (SVMs) and Logistic Regression (LogReg).

SVMs and LogReg are two widely used algorithms for classification tasks. SVMs are well-suited for handling non-linear data by finding the optimal linear boundary to separate the classes. Furthermore, they can be applied to both linear and non-linear classification problems. On the other hand, LogReg is known for its simplicity and efficiency, making it a practical choice for many classification problems. Additionally, it offers a probabilistic output, making it useful for predicting the likelihood of a particular class. We selected these methods for our classification problem due to their versatility, ease of implementation, and capability to deliver high-accuracy results.

To assess the performance of our classifiers, we use two metrics: accuracy (Acc) and false discovery rate (FDR). Accuracy is a measure of the overall performance of the classifier.
Accuracy is calculated by dividing the number of correct predictions the classifier makes by the total number of samples. The formula for accuracy is:
\begin{equation}
\text{Acc (\%)} = \frac{\text{TP} + \text{TN}}{\text{TP} + \text{TN} + \text{FP} + \text{FN}} \times 100
\end{equation}
Here, TP, TN, FP, and FN represent the number of true positive, true negative, false positive, and false negative predictions, respectively.
The False Discovery Rate (FDR) indicates the proportion of false positive predictions. It can be calculated using the following formula:
\begin{equation}
\text{FDR (\%)} = \frac{\text{FP}}{\text{TP} + \text{FP}} \times 100
\end{equation}
By using these two metrics, we can evaluate the performance of our classifiers in terms of both overall accuracy and the rate of false positive predictions.
\section{Results}
We utilized the Classification learner toolbox in MATLAB \cite{mathworks} to train and evaluate both a linear kernel Support Vector Machine (SVM) and a Logistic Regression (LogReg) classifier. The toolbox employed Bayesian optimization to optimize the hyperparameters of both models. Our results showed that both classifiers performed similarly in the training phase, with LogReg having an accuracy of 95.3\% and SVM with 96.4\%. However,  SVM slightly outperformed LogReg.The confusion matrices for the training process are displayed in TABLE.~\ref{tab:conf}.  Moreover, both classifiers had a satisfactory rate of false positives and false negatives.\\
\begin{table}[!h]
\renewcommand*{\arraystretch}{1.1}
\begin{subtable}[c]{0.49\linewidth}
    \centering
   \begin{tabular}{lrcc}
    \multicolumn{2}{c}{}&\multicolumn{2}{c}{Predicted class}\\
    \multicolumn{2}{c}{}&0 &  1\\
    \cline{3-4}
     \parbox[c]{-2mm}{\multirow{-1.5}{*}{\rotatebox[origin=c]{90}{True Class}}} & 0 & \multicolumn{1}{|c|}{95.1\%} & \multicolumn{1}{|c|}{4.9\% }\\
    \cline{3-4}
    & 1 & \multicolumn{1}{|c|}{4.5\%} & \multicolumn{1}{|c|}{95.5\%}\\
    \cline{3-4}
    \end{tabular} 
\subcaption{LogReg}
\end{subtable}
\begin{subtable}[c]{0.49\linewidth}
   \begin{tabular}{lrcc}
    \multicolumn{2}{c}{}&\multicolumn{2}{c}{Predicted class}\\
    \multicolumn{2}{c}{}&0 &  1\\
    \cline{3-4}
     \parbox[c]{-2mm}{\multirow{-1.5}{*}{\rotatebox[origin=c]{90}{True Class}}} & 0 & \multicolumn{1}{|c|}{96.2\%} & \multicolumn{1}{|c|}{3.8\% }\\
    \cline{3-4}
    & 1 & \multicolumn{1}{|c|}{3.5\%} & \multicolumn{1}{|c|}{96.5\%}\\
    \cline{3-4}
    \end{tabular} 
\subcaption{SVM}
\end{subtable}
\caption{Trainnig confusion matrices for LogReg and SVM. The results demonstrate a good accuracy for both classifiers during the training phase.}
\label{tab:conf}
\end{table}

Furthermore, during the application of disturbances, there may be short time intervals where the object is stationary. The contacts can be considered stable but mistakenly labeled unstable, leading to incorrect labeling. Additionally, when the fingers close on the object, one finger may reach it first, apply a force that attains the threshold, and stop, resulting in stable contact. However, when another finger reaches the object and pushes in the opposite direction, it creates changes in the measured force and leads to misclassification, resulting in the data being labeled as unstable contact.

We investigated the influence of $N_w$, the size of the temporal window used in the DWT decomposition, on the performance of the Logistic Regression (LogReg) classifier. Results showed that the accuracy of the classifier improved as the window size increased TABLE.~\ref{tab:nwtab}. However, beyond $N_w=14$, the accuracy (Acc), False Negative (FN), and False Positive (FP) rates remained relatively unchanged. 
We opted not to explore values of $N_w$ above 18, as a larger temporal window could introduce interference from past instability events on the current state of the contact. 
As a result, for the rest of this work, we used $N_w=14$.
\begin{table}[!ht]
\begin{center}
\begin{tabular}{lccccccccc}
\hline
$N_w$ & 4 & 6 & 8 & 10 & 12 & 14 & 16 & 18\\ \hline
FPR & 8,6 & 7,2 & 6,4 & 5,8 & 5,5 & 5,2 & 5,0 & 4,9\\ 
FNR & 5,3 & 5,4 & 5,3 & 5,0 & 4,8 & 4,7 & 4,5 & 4,5\\ 
FDR & 8,6 & 7,3 & 6,5 & 6,0 & 5,6 & 5,4 & 5,1 & 5,1\\ 
Acc & 93,0 & 93,7 & 94,2 & 94,6 & 94,9 & 95,0 & 95,2 & 95,3 \\ \hline
\end{tabular}
\end{center}
\caption{LogReg perforrmance (\%) versus $N_w$. This table compares the classifier's performance using different temporal window sizes $N_w$ used to extract features.}
\label{tab:nwtab}
\end{table}

We conducted a feature ablation study to evaluate each feature's contribution to the LogReg classifier's performance. As seen in TABLE.~\ref{tab:feat_abl}, the results indicate that the standard deviation $\sigma$ and force amplitude $F_a$ were the most significant contributors to the classifier's performance. We also observed that the mean $m$ and $F_a$ were highly correlated, which makes sense as $m$ can be viewed as a filtered version of $F_a$. However, the classifier appeared to perform slightly better using $F_a$ instead of $m$. Therefore, we removed $m$ from the feature vector $\Phi$ for the rest of this work. The results also demonstrate the effectiveness of using the $\sigma$ feature. When $\sigma$ is removed from the features vector $\Phi$, the accuracy decreases by $14.0\%$, and the FDR increases from $5.1\%$ to $20.3\%$.\\
\begin{table}[!ht]
\begin{center}
\begin{tabular}{cccccrrrr}
\hline
\multicolumn{4}{l}{Features}   &{}  & \multicolumn{4}{l}{Performance (\%)}  \\ 
\cline{1-4} \cline{6-9}  
$F_a$ & $F_{tip}$ & $m$ & $\sigma$ &{}& FPR  & FNR & FDR & Acc   \\
\cline{1-4} \cline{6-9} 
\mC & \mC & \mC & \mC &{} & 4.9 & 4.5 & 5.1 & 95.3 \\
\mX & \mC & \mC & \mC &{} & 4,9 & 4,6 & 5,0 & 95,3 \\
\mC & \mX & \mC & \mC &{} & 7,0 & 6,7 & 7,2 & 93,1 \\
\mC & \mC & \mX & \mC &{} & 4,9 & 4,6 & 5,0 & 95,3 \\
\mC & \mC & \mC & \mX &{} & 20,5 & 16,9 & 20,3 & 81,3 \\
\mX & \mX & \mC & \mC &{} & 7,1 & 6,7 & 7,3 & 93,1 \\
\mC & \mX & \mX & \mC &{} & 7,0 & 6,7 & 7,2 & 93,2 \\
\mC & \mC & \mX & \mX &{} & 20,9 & 16,8 & 20,5 & 81,1 \\
\mX & \mC & \mX & \mC &{} & 5,2 & 4,7 & 5,4 & 95,0 \\
\mX & \mC & \mC & \mX &{} & 20,3 & 16,8 & 20,2 & 81,4 \\
\mC & \mX & \mC & \mX &{} & 26,1 & 32,9 & 28,6 & 70,6 \\
\hline
\end{tabular}
\caption{Performance of the LogReg Classifier in feature ablation study. This table shows the accuracy of the classifier with different features removed (\mX) from the input data. The results provide insights into which features among $F_a$, $F_{tip}$, $m$, and $\sigma$ are most relevant for the contact type classification.}
\label{tab:feat_abl}
\end{center}
\end{table}

\begin{figure}[b!]
    \centering
    \includegraphics[width=0.90\linewidth]{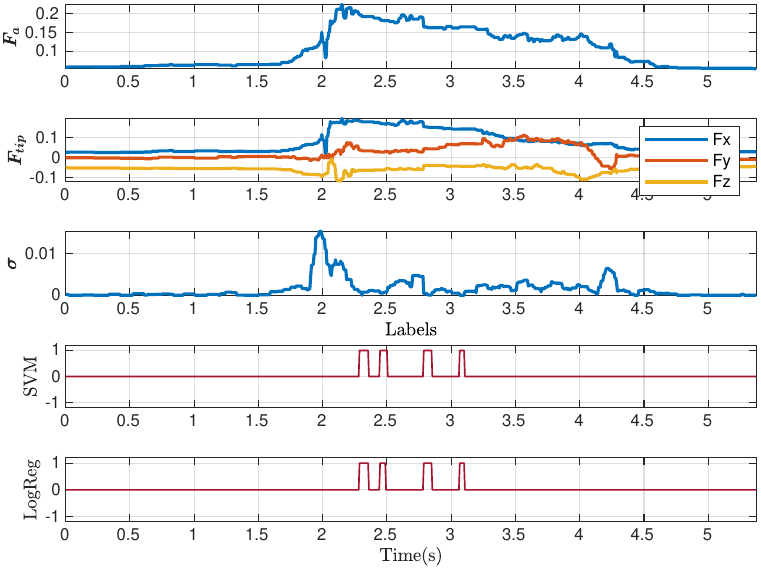}
    \caption{Experiment 1: Failed grasp. The predicted labels for failed grasps, where the hand could not securely grasp the object. The contact was maintained in small periods. the contact type was successfully classified as unstable by both SVM and LogReg.}
    \label{fig:exp1}
\end{figure}
\begin{figure}[b!]
    \centering
    \includegraphics[width=0.9\linewidth]{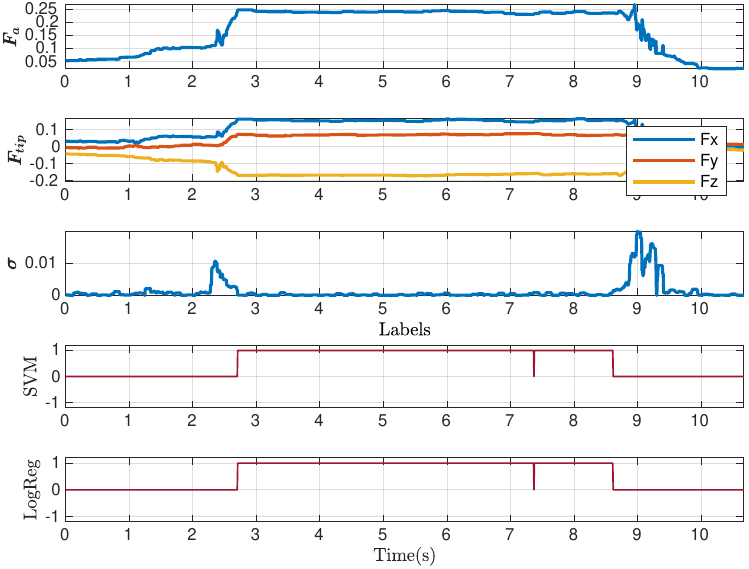}
    \caption{Experiment 2: Successful grasp with disturbances. The predicted labels for an experiment were where the hand gradually closed on the object, secured its grip, and then released the object. }
    \label{fig:exp2}
\end{figure}

Now that we designed our classifier, we evaluate it using previously unseen data.
This simulates how the classifier works in an online manner.
The performance for both classifiers was satisfactory, with $95.64\%$ for LogReg and $96.2\%$ for SVM.
Fig.~\ref{fig:exp1} and Fig.~\ref{fig:exp2} show the predicted labels for two experiments.
In
\textbf{Experiment 1} (Fig.~\ref{fig:exp1}), the recorded data represents a failed grasp where the hand closed on an object but failed to stabilize it, resulting in slippage and contact loss. Both LogReg and SVM were able to classify this data accurately.
In \textbf{Experiment 2} (Fig.~\ref{fig:exp2}), the hand closed on an object and locked it before we started applying disturbances.
The force amplitude $F_a$ increased, and the standard deviation ($\sigma$) decreased as the finger pushed to lock the object. Before locking, the finger may have pushed in opposite directions, causing the standard deviation to increase and the predicted class to change between $0$ and $1$. After locking the object, the variation in force amplitude and standard deviation were minimal. However, some glitches in the fingers were observed due to the use of a simple PD controller for the Allegro hand. Both classifiers could detect these abrupt changes in force and classify them as unstable contact. These results demonstrate that the classifiers can differentiate between stable and unstable contact, especially when disturbances or slippage occur, which is critical in determining the stability of the grasp.
Furthermore, the LogReg classifier was also trained for the rest of the finger.
It gave a similar performance to the previous classifier. For the middle finger, Acc $95.8\%$, FDR $12.6\%$, and for the thumb: Acc $95.7\%$, FDR $3.5\%$. Due to the effectiveness and simplicity of the logistic regression, we chose it over SVM; we implemented it with the robot. We attached a video to show the performance of these classifiers in real-time. In conclusion, the results show the effectiveness of the chosen method.
 \section{Discussion}
In this part, we will compare our work to the most relevant studies in the field of slip and instability detection for robotic hands.
Zhang et al. \cite{zhang2016initial} present a dynamical model that considers disturbances during the initial grasp.
In practice, they use the force-sensitive tactile sensors (FSR) to measure the grasping force and then apply the Haar wavelet transform to extract detail coefficients. 
Moreover, they consider an object slipping when the energy of the details exceeds a threshold determined through experimentation. 
We implemented and tested their method on $F_x$, i.e., the normal component of $F_{tip}$. We used SVM to find the optimal threshold. The accuracy was not more than $62.7\%$, and the FDR was up to $49.1\%$. The performance of our method drastically exceeds theirs (Acc $95.3\%$, FDR $5.1\%$).
While this study provides valuable insights, it is limited by the adoption of different thresholds for slips caused by gravity and disturbance. In addition, using energy values only for slip prediction can result in errors in real-world applications. 
In contrast, in our work, we also measure the amplitude of the force, its three components, and the standard deviation. 
The amplitude ($F_a$) provides information about the magnitude of the grasping force.
This measure is crucial for classification because energy variation with large grasping forces is relatively higher than with small grasping forces, even for slips with the same velocity. 
Shear forces also carry information that contributes to the detection of slips. Hence, three components of the grasping force are used in our work. 
Finally, we argue that the standard deviation ($\sigma$) is a better measure for abrupt changes in the detail coefficients than energy. While both energy and $\sigma$ measure data dispersion, $\sigma$ takes the mean of the details into account. In the presence of noise, the energy will amplify the values of the details and lead to false detection of slip, whereas the standard deviation will stay relatively small. 
Overall, our method is more effective since it relies on richer features and employs an automatic learning algorithm for slip detection, i.e., finding a nonlinear decision boundary rather than a hand-tuned threshold.

The second study we consider is James et al.~\cite{james2020slip}, in which a multi-fingered hand is used with optical tactile sensors. A global classifier for the whole hand and local classifiers for each fingertip were trained over collected tactile data. They used different techniques, including SVM and LogReg, to classify ``slip'' and ``static'' data. In the training phase, their results are similar to ours, i.e., their global classifier reaches an accuracy of 96\% for SVM and 95.7\%  for LogReg. In our method, SVM and LogReg scored 96.0\% and 95.6\%, respectively. Their local classifier had a relatively low performance compared to the global classifier, with a maximum accuracy of 83.9\% for SVM. In online training, the global classifier got an accuracy of 93.75\% 
The performance of our classifiers surpasses the performance of local classifiers.
Furthermore, their study only considered slippage due to gravity, i.e., when the object moves downward on a vertical axis.
However, in this work, we considered slippage due to gravity and external forces in all possible directions.

Grover et al.~\cite{grover2022learning} used barometric tactile sensors to collect tactile data in different scenarios. 
They then trained a classifier using a convolutional neural network to classify ``slip'' and ``non-slip''. The classifier scored 91.4\% in the training phase and 87.5\% in the testing phase (our LogReg classifier scored 95.2\% in the training and 95.6\% in the online testing).
As for the previous work by James et al.~\cite{james2020slip}, their work only predicts slip without providing relevant features that can be exploited in the control loop.
In conclusion, our work presents a more robust and effective approach to slip and instability detection for robotic hands compared to previous studies. 
By incorporating proposed features into the control loop, our approach has the potential to achieve reliable performance in a wider range of real-world applications.\\
\section{ Conclusion }
We proposed a new model-free method for predicting slippage and disturbances in grasping and manipulation tasks using 3-axial tactile sensors in this work.
The method is based on collecting tactile data and extracting relevant features to train and evaluate two classification methods, i.e., support vector machine and logistic regression. 
The performance of the classifiers was evaluated with different sizes of the time window and through feature ablation.
The final results show that the logistic regression classifier accurately detects instability caused by slippage and disturbances.
Using DWT for feature extraction and a logistic model for regression makes our classifier an efficient tool for real-world implementations. 
Furthermore, our method provides the type of contact (``stable'' vs. ``unstable'') and a measure of instability ($\sigma$) for each fingertip separately in addition to the probabilistic output of Logistic regression.
This makes it suitable for independent feedback control strategies for grasp adaptation and optimization.

The labeling process can be considered a limitation of our work. 
To independently assess the stability of each fingertip, we opted for manual labeling of the data.
Unfortunately, this process is prone to human errors and can be time-consuming. A more robust and efficient labeling method can be implemented in future work by placing markers on the fingertips and the object and tracking their relative movements. Moreover, we consider incorporating the features calculated in our study into the control loop. These features provide important information about disturbances and slippage's direction, amplitude, and stability. Our results show that the standard deviation decreases as contact stability is reached, indicating that the grasp is becoming more secure. By utilizing this information, future work can focus on adapting and optimizing the grasp online.

\section*{Acknowledgment}
Thanks to Alexander Schmitz and Matthias Kuus From Xela Robotics for their support and assistance with the tactile sensors.

\addtolength{\textheight}{-12cm}   


\bibliographystyle{IEEEtran}
\bibliography{IEEEabrv,references}

\end{document}